\documentclass[journal]{IEEEtran}
\usepackage{datetime}
\usepackage{times}
\usepackage{epsfig,graphicx}  
\usepackage{graphicx, amssymb}
\usepackage{amsfonts, enumerate}
\usepackage{latexsym,subfigure,makeidx}
\usepackage{url}
\usepackage{tabls}
\usepackage{algorithm}
\usepackage{algpseudocode}
\usepackage{lscape}
\usepackage[sort,space]{cite}
\usepackage{pifont}
\usepackage{tabularx}
\usepackage{rotating}
\usepackage{hhline}
\usepackage{textcomp,booktabs}
\usepackage{xcolor}
\usepackage{array}
\usepackage{amsmath,bm}
\usepackage{caption}
\usepackage{multirow}
\usepackage{verbatim}
\usepackage{amsthm}
\usepackage{framed}
\usepackage{diagbox}

\ifCLASSINFOpdf
   \graphicspath{{./}}
\else
\fi
\ifCLASSOPTIONcompsoc
  \usepackage[caption=false,font=normalsize,labelfont=sf,textfont=sf]{subfig}
\else
 \usepackage[caption=false,font=footnotesize]{subfig}
\fi
\def\NoNumber#1{{\def\alglinenumber##1{}\State #1}\addtocounter{ALG@line}{-1}}
\begin{document}
%
\title{ Edge Intelligence for Energy-efficient Computation Offloading and Resource Allocation in 5G Beyond}
%
%

%
\author{Yueyue~Dai, \IEEEmembership{Member,~IEEE},
         Ke Zhang,
         Sabita~Maharjan, \IEEEmembership{Senior~Member,~IEEE},
        and~Yan~Zhang, \IEEEmembership{Fellow,~IEEE}
\thanks{ Copyright (c) 2015 IEEE. Personal use of this material is permitted. However, permission to use this material for any other purposes must be obtained from the IEEE by sending a request to pubs-permissions@ieee.org.}
\thanks{This research is partially supported by the National Natural Science Foundation of China under Grant 61941102 and  partially supported by the Xi’an Key Laboratory of Mobile Edge Computing and Security, under Grant No. 201805052ZD3CG36.}
\thanks{Y. Dai and K. Zhang are  with the School of Information and Communication Engineering, University of Electronic Science and Technology of China, Chengdu 611731, China (email:yueyuedai@ieee.org; zhangke@uestc.edu.cn).}
\thanks{S. Maharjan and Y.  Zhang (corresponding author) are with Department of Informatics, University of Oslo, Norway, and also with Simula Metropolitan Center for Digital Engineering, Norway. (email: sabita@ifi.uio.no, yanzhang@ieee.org).}
}



\maketitle

	\begin{abstract} 5G beyond  is an end-edge-cloud orchestrated network that can exploit heterogeneous capabilities of the end devices, edge servers, and the cloud and thus has the potential to  enable computation-intensive and delay-sensitive applications via computation offloading. However, in multi user wireless networks, diverse application requirements and the possibility of various radio access modes for communication among devices make it challenging to design an optimal computation offloading scheme.  In addition, having access to complete network information that includes variables such as wireless channel state, and available bandwidth and computation resources, is a major issue.
	  Deep Reinforcement Learning (DRL) is an emerging technique to address such an issue with limited and less accurate  network information. In this paper, we utilize DRL  to design an optimal computation offloading and resource allocation strategy for minimizing system energy consumption. We first present a  multi-user end-edge-cloud  orchestrated network where all devices and base stations have computation capabilities. Then, we formulate  the joint computation offloading and resource allocation problem as a Markov Decision Process (MDP)  and propose  a new DRL algorithm to minimize system energy consumption. Numerical results  based on a real-world dataset demonstrate that the proposed  DRL-based algorithm  significantly outperforms the benchmark policies  in terms of system energy consumption. Extensive simulations show that learning rate, discount factor, and number of devices have considerable influence on  the performance of the proposed  algorithm. 

	\end{abstract}

\begin{IEEEkeywords}
Deep reinforcement learning,  Edge computing, Computation offloading, 5G Beyond.
\end{IEEEkeywords}

%

\IEEEpeerreviewmaketitle
\section{Introduction}
Recent advancements in  Internet of Things (IoT) and 5G wireless networks have paved a way towards realizing new application such as surveillance, augmented/virtual reality, and face recognition, which usually are both computation and caching resource intensive.  For battery-powered  and resource-constrained  devices, such as wearable devices, on-device sensors, and smart phones,  it is therefore challenging to support such applications, in particular, delay sensitive applications  \cite{abbas2018mobile}.     To alleviate computation limitations and to prolong battery lifetime  of IoT devices,   Mobile Cloud Computing (MCC) leverages offloading  computation intensive tasks to the centralized  cloud server.  MCC however may result in high communication latency, low coverage, and lagged data transmission thus limiting its applicability for real-time applications such as  automated driving and smart navigation. Edge computing is a promising paradigm to address such issues associated with MCC,  by pushing additional computation resources available at the edge servers \cite{mach2017mobile}.

Utilizing  computation offloading, IoT devices can offload their computation-intensive and delay-sensitive  tasks to nearby  distributed base stations, access points, and road-side units, integrated with edge servers \cite{rodrigues2018cloudlets}. There has been considerable amount of work focusing on  computation offloading.  The authors  in  \cite{zhao2019computation} proposed to offload computation tasks  to  lightweight and distributed vehicular edge servers to minimize task processing latency.  In \cite{you2017energy} and \cite{8611210}, the authors proposed to offload computation tasks to the  Macro-cell Base Stations (MBS) for minimizing the maximum weighted energy consumption.  The authors in \cite{dai2018joint} proposed to offload computation tasks to multiple  distributed Small-cell Base Stations (SBS) and the MBS,  to minimize energy consumption.   However, the computation capacities of these ubiquitous edge servers are often limited such that offloading all devices' tasks to edge servers  may instead result in a higher computation latency. { Moreover,  computation offloading especially in  ultra dense and heterogeneous 5G networks can lead to a higher  task processing delay compared to local  computing in devices, due to additional delay caused by the need to transmit the data to base stations.  Local computing, on the other hand, can  result in  low task execution latency  since in that case there is no delay added due to transmitting the task and the result of task execution back and forth.  }

To  make full use of   heterogeneous computation capabilities of devices, edge servers, and the cloud,  5G beyond networks offers  an end-edge-cloud  orchestrated architecture that can concurrently support local computing and computation offloading. Optimizing computation offloading is this case means deciding whether to execute tasks locally at the end devices or offload them to edge servers or to offload a certain amount of them  to the central server. However, due to the highly dynamic and stochastic nature of wireless systems,  the lack of complete network information such as  wireless channel state, available bandwidth and computation resources, makes it challenging to design  an optimal offloading strategy. In addition, diverse application-specific requirements and the use of multiple radio access modes, make the problem sufficiently complex.

Edge intelligence is a promising paradigm that leverages capability at the network edge  to enable real-time services \cite{lu2020edge},  \cite{yueyuedai}. By integrating AI into edge networks, edge intelligent servers have full insight of working environments in terms of the resource correlation between heterogeneous types and the feasibility of cooperation with adjacent nodes \cite{rodrigues2019machine}. Moreover, as edge services typically utilize edge resources to estimate the environment, edge resource scheduling does not require the information of the entire network from  the central cloud. Transmitting locally learned information to the distributed AI processing entities is more efficient than obtaining resource scheduling strategies from remote central AI nodes \cite{zhang2019edge}. Thus, diverse servers at the network edge can  incorporate heterogeneous resources and offer flexible communication and computing services in an efficient manner.

In this work, we propose an edge intelligence  empowered 5G beyond  network by jointly optimizing computation offloading and scheduling the distributed resources. We consider an end-edge-cloud  orchestrated network consisting of  devices, edge computing servers and a cloud server where computing tasks can be executed locally with cooperation between the edge computing servers and the cloud server.   Due to highly dynamic of wireless environment and the diverse application requirements, we exploit  Deep Reinforcement Learning (DRL)  to learn dynamic network topology and time-varying wireless channel condition and then design a joint computation offloading and resource allocation scheme. The key contributions of our work are summarized as follows:

\begin{itemize}
\item  We present a multi-user end-edge-cloud orchestrated network  where devices can offload computation-intensive and delay-sensitive tasks to a particular edge server on an SBS and/or the MBS, respectively.

\item We formulate the joint computation offloading and resource allocation problem to minimize system energy consumption taking stringent delay constraints into account.

\item We propose  a DRL-based computation offloading  and resource allocation  algorithm to reduce energy consumption. Numerical results based on a real-world dataset  demonstrate that the proposed  DRL-based algorithm  significantly outperforms the benchmark policies. 
\end{itemize}

The remainder of this paper is organized as follows. The related work is discussed in Section \ref{rw}. We then present the system model  of  the end-edge-cloud orchestrated network in Section  \ref{sm}.  The joint computation offloading and resource allocation problem is formulated in Section \ref{pF}. We design a  DRL- based algorithm to solve the proposed computation offloading problem in Section \ref{DRL}. Numerical results  are presented in Section \ref{sr}. Finally, the paper is concluded in Section \ref{c}.

 \section{Related work}
 \label{rw}
 
 In this section, we first introduce the state-of-the-art  researches on computation offloading in wireless networks and then discuss the work that exploits DRL for computation offloading. 
 
 \subsection{Computation Offloading in Wireless Networks}
 In 5G beyond, devices are connected to the infrastructure  through wireless communication.  This introduces high agility in resource deployment and facilitates pushing edge services closer to the end users. By offloading tasks directly to edge servers, the resource burden of the devices can be alleviated, and the application processing latency can also be considerably reduced.
 
 {Computation offloading in wireless networks is widely investigated.} For instance, the  authors in \cite{guo2018mobile} considered an  ultra-dense network and proposed a two-tier greedy  offloading scheme to minimize computation overhead. The authors in \cite{rodrigues2016hybrid} proposed a computation offloading scheme to determine whether the computation-intensive tasks should be executed  locally or remotely on the MEC servers. The authors in \cite{you2017energy} considered  a multi-user  system based on time-division multiple access and orthogonal frequency-division multiple access, and formulated an computation offloading optimization problem to minimize the weighted  energy consumption under the constraint on computation latency. {In \cite{zhao2019novel}, the authors proposed  a three-hierarchical offloading optimization strategy, which jointly optimizes  bandwidth, offloading strategy to reduce the system latency and energy consumption.}  In \cite{wang2018multi}, the authors proposed a multi-user MEC system by integrating  a multi-antenna Access Point (AP)  with an MEC server such that mobile users  can execute their respective tasks locally or offload them to the AP.   The authors in \cite{dai2018joint}  proposed  a computation offloading algorithm by   jointly considering the offloading decision, computation resources, and the transmit power for computation offloading. The authors in \cite{samanta2019adaptive} proposed an iterative computation offloading  algorithm to  improve the resource utilization efficiency in wireless networks. 
 
 The optimization algorithms of the above researches often require complete and accurate network information. However,  the wireless network is time-varying such that  the  accurate network information of wireless channel state, bandwidth and computation resources are  difficult to obtain.  Thus, the above optimization algorithms may not be practical in the wireless networks.

 \subsection{Deep Reinforcement Learning for Computation Offloading }

 Deep reinforcement learning is a promising technique  to address the problems with uncertain and stochastic feature and it also can be utilized to solve the sophisticated optimization problem \cite{ahmed2019deep},\cite{dai2020deep}, \cite{lu2020blockchain}.   State-of-the-art offers a good volume of recent works that have utilized DRL for optimizing computation offloading in wireless networks. For instance, the authors in \cite{wang2019smart} proposed  a smart DRL-based resource allocation scheme in wireless networks, to allocate the computing and network resources for reducing service time and balancing resources. The authors in \cite{huang2019deep} formulated the  joint task offloading decision and bandwidth allocation optimization problem and proposed a  Deep-Q Network (DQN) based algorithm to solve the optimization problem.  {The authors in \cite{yang2020lessons} proposed an edge learning-based offloading algorithm where deep learning tasks  are defined as computation tasks and they can be offloaded to edge servers to improve inference accuracy. } The authors in \cite{zhang2019deep} proposed  a  deep Q-learning based task offloading scheme to select the optimal edge  server  and the optimal transmission mode to maximize task offloading utility. The authors  in \cite{8690980}  proposed  a  deep Q-network based  computation offloading scheme to minimize the long-term cost. The authors in \cite{xie2019backscatter} proposed the Double Deep Q-Network (DDQN) based  backscatter-aided hybrid data offloading scheme to  reduce  power consumption in data transmission. The authors in \cite{lu2019blockchain} proposed to utilize federated learning to design a distributed data offloading and sharing scheme for a secure network. The above works only decide whether to execute tasks locally or offload to edge servers. However, for an end-edge-cloud network,  computation offloading is a  three-tier offloading decision-making problem, and  the heterogeneous computation capabilities of devices, edge servers, and the cloud add the complexity of the resource allocation. Thus, it is quite difficult to  find the optimal solution of the joint three-tier computation offloading and resource allocation problem in end-edge-cloud networks. On the other hand, the above works are mainly based on discrete  action space, and thus are of limited value for the problem with continuous action spaces. 
 
 In this paper, we consider an end-edge-cloud orchestrated  network consisting of mobile devices, edge computing servers and a cloud server, {to  jointly optimize computation offloading and scheduling heterogeneous computation resources.  } Since the action of three-tier offloading decision-making is  discrete  and the action of heterogeneous computation resource allocation is continuous, we incorporate action refinement in DRL and design a new DRL algorithm to concurrently optimize computation offloading and resources allocation.
 
\section{System Model}
\label{sm}

\begin{figure}
	\centering
	\includegraphics[width =3.5 in]{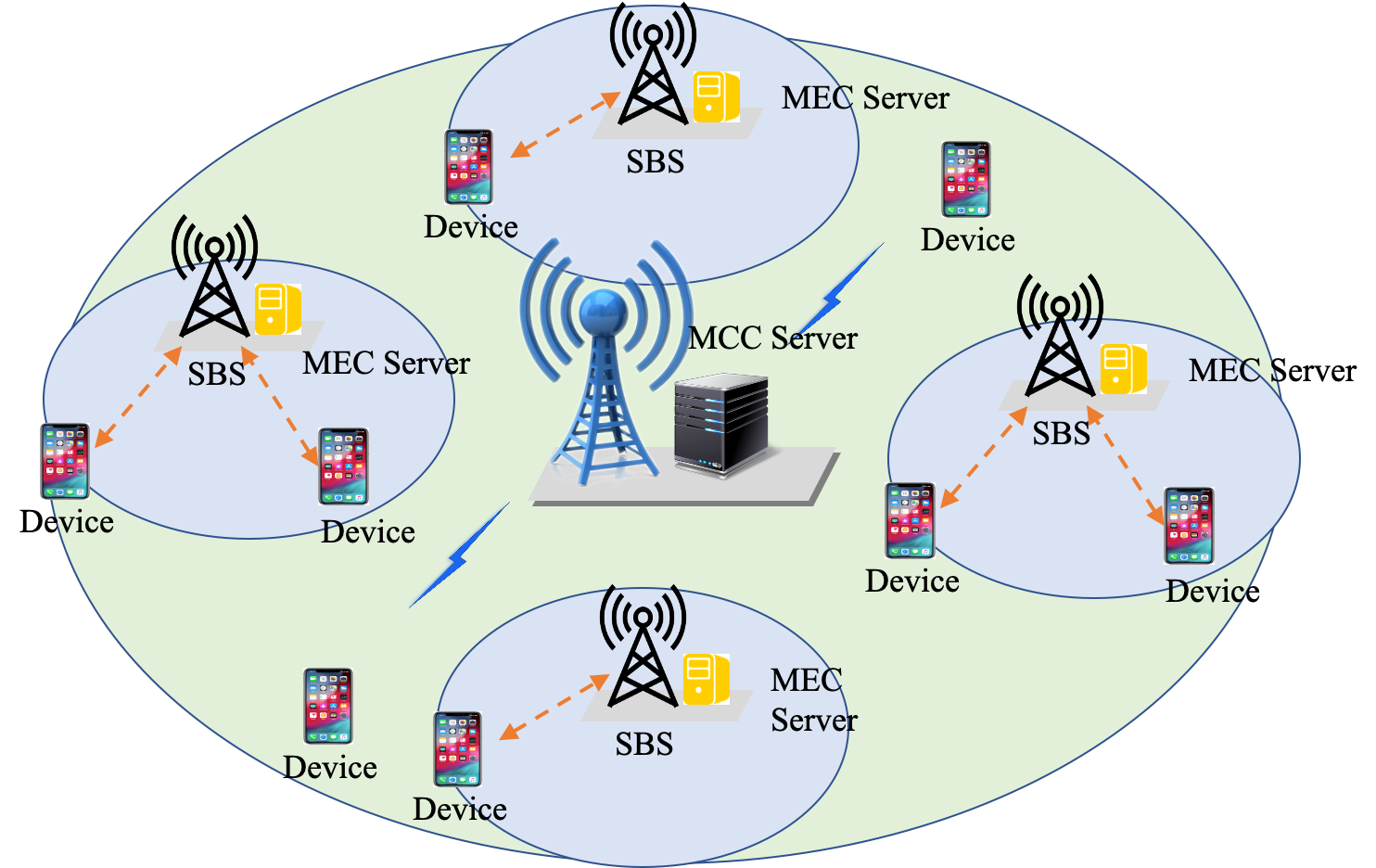}
	\caption{The scenario of end-edge-cloud  heterogeneous networks with multiple users .}
	\label{overview}
\end{figure}
We consider  a multi-user end-edge-cloud orchestrated  5G heterogeneous network, which  consists of one  MBS and $M$ SBSs, as  shown in Fig. \ref{overview}.  Let  $\mathcal{B} = \{b_0,b_1,...,b_M\}$ denote the set of all base stations, where $b_0$ is the MBS.   There are  $N$  devices randomly distributed under the coverage of the MBS. The set of devices  is denoted as $\mathcal{U}=\{u_1,..,u_N\}$.  Each SBS is equipped with an edge server to provide computation  resource close to the end devices. The MBS is equipped with a cloud server which has relatively more computation resource. Each device has a  computation-intensive and delay-sensitive task to be executed within a certain execution deadline, for applications such as  augmented/virtual reality, online interactive game, and navigation. Let us denote the task as  $D_i=(d_i,c_i,\tau_i)$, where $d_i$ and  $c_i$  denote the input data size and the required computation resource, respectively, and  $\tau_i$ indicates the upper limit on latency for completing the task.

Devices may compute their tasks locally if the task completion delay will be within the limit. Otherwise, devices will offload their tasks to nearby edge servers for edge computing. We consider devices are able to offload their tasks to a nearby SBS or to the MBS. Let $x_i^l,y_{ij}^s,z_i^m \in\{0,1\}$ denote:  task $ D_i$  is computed locally,  task  $ D_i$  is offloaded to SBS $j$, and task $ D_i$ is  executed at the MBS, respectively.  { Each device only can choose one way to execute its task, thus offloading variables should  satisfy the constraint $x_i^l+\sum_{j=1}^My_{ij}^s+z_i^m=1$. For convenience, the key notations of our system model are summarized in Table  \ref{table_1}.

\begin{table}[!t]
	\renewcommand{\arraystretch}{1.3}
	\caption{ Summary of Key Notations}
	\label{table_1}
	\begin{tabular}{|l|l|}
		\hline
		\bfseries Notation& \bfseries Definition\\
		\hline
		$\mathcal{B}$ &The set of base stations\\
		$\mathcal{U}$   &The set of  devices\\
		$D_i=(d_i,c_i,\tau_i)$ &Computation task of device $i$ where  \\& $d_i$, $c_i$ and  $\tau_i$ are input data size,   \\& required computation resource,  \\& and  the upper limit on latency  \\& for completing the task, respectively \\		
		$x_i^l,y_{ij}^s,z_i^m $ &Offloading variables where $x_i^l$ means \\&local computing, $y_{ij}^s$ means offloading \\& to SBS, $z_i^m $ means offloading to MBS\\
		$R_{ij}^s, R_{i0}^m$ 	&Wireless communication data rate  between \\& devices and base stations\\
		$p_i$  &Transmission power of  device $i$\\
		$h_{ij}^s, h_{i0}^m$ & Wireless channel gain between \\& devices and base stations\\
		$\sigma^2$ &Noise power\\
		$b_s,b_m$ &  Bandwidth of SBS and MBS, respectively \\	
		$r_{ij}^s, r_{i0}^m$ 	&Distance  between devices and base stations\\
		$T_i^l (E_i^l)$ &Task computing time (Energy consumption) of  \\& local computing\\
		$T_{ij}^s (E_{ij}^s)$ &Task completion time (Energy consumption) for \\& offloading task to SBS\\
		$T_{i0}^m (E_{i0}^s)$ &Task completion time (Energy consumption) for \\& offloading task to MBS\\
     	$f_i^l,F_j^s,F_0^m$ &The total computation resource of device $i$,\\& SBS $j$, and the MBS, respectively. \\
		$f_{ij}^s,f_{i0}^m$& Variables of computation resource that $f_{ij}^s$ is the \\&  computation resource of SBS  $j$ assigned to device \\&  $i$ and 	$f_{i0}$  is the computation resource of MBS  \\&   assigned to device $i$\\
		 $e_j, e_0$ &The energy consumption per CPU cycle of \\&SBS $j$ and the MBS, respectively. \\
		\hline
	\end{tabular} 
\end{table}
}

\subsection{Communication Model}
Devices are associated with  the MBS or the SBSs. The wireless communication data rates between devices and base stations 
can vary depending on  transmission power, interference, bandwidth.   As shown in Fig. \ref{overview}, SBSs are deployed at hot spots and they do not have significant coverage overlap. To utilize spectrum efficiently, all SBSs reuse the MBS's frequency resource. Orthogonal Frequency Division Multiple Access (OFDMA)  is adopted in wireless networks such that the interference between the MBS and SBSs can be well suppressed \cite{8399837} .


 In this work, we consider devices first communicate with  the nearest SBS to perform computation offloading. We define $\gamma_j^s$ as the coverage radius of SBS $j$. If the distance between  device $i$ and SBS $j$ is less than $\gamma_j^s$ (i.e.,  $r_{ij}^s<\gamma_j^s$), device $i$  can  communicate with  SBS $i$.    The  wireless communication data rate  between  device $i$ and  SBS $j$ can be expressed as 
 \begin{equation}
 \label{rate_sbs}
 R_{ij}^s ={b^s}\log_{2}(1+\frac{p_ih_{ij}^s(r_{ij}^s)^{-\alpha}}{\sigma^2+\sum_{i'\in\mathcal{U}/\{i\},j'\in\mathcal{B}/\{j\}}{p_{i'}h_{i'j'}(r_{i'j'}^s)^{-\alpha}}}), 
 \end{equation}
 where $b^s$ denotes channel bandwidth between  device $i$ and SBS $j$, $p_i$ is the transmission power of device $i$, $h_{ij}^s$ is the channel gain  between  device $i$ and SBS $j$,  $\alpha$ is the path loss exponent, and $\sigma^2$ is the noise power.  $ \sum_{i'\in\mathcal{U}/\{i\},j'\in\mathcal{B}/\{j\}}{p_{i'}h_{i'j'}(r_{i'j'}^s)^{-\alpha}}$ is the interference from  other SBSs.
 
 If  device $i$ does not lie within the coverage of any SBS, it will communicate  with the MBS. The  wireless communication data rate between  device $i$ and the MBS can be defined as 
\begin{equation}
\label{rate_mbs}
R_{i0}^m ={b^m}\log_{2}(1+\frac{p_ih_{i0}^m(r_{i0}^m)^{-\alpha}}{\sigma^2}), 
\end{equation}
where $b^m$ represents the channel bandwidth between  device $i$ and the MBS, $h_{i0}^m$ is the channel gain  between  device $i$ and the MBS, $r_{i0}^m$ is the distance between  device $i$ and the MBS.

\subsection{Computation Model}
For device $i$,  task $D_i$ can be executed in three ways, i.e., executed  locally on the device, offloaded to an SBS, or offloaded to the MBS. 

 \textit{1) Local Computing}: If device $i$ chooses to compute  task $D_i$ locally, the whole computation resource of this device will be utilized to compute the task.  We denote $f_i^l$ as the computation  resource (i.e., CPU cycles per second) of device $i$. The local  task completion delay only includes  the task computing delay, which can be expressed as
 \begin{equation}
 \label{local_time}
 T_i^l = {c_i}/{f_i^l},
 \end{equation}

The energy consumption of unit computation resource is $\varsigma  ({f_i^{l}})^2$, where $ \varsigma $ is the effective switched capacitance depending on the chip architecture
\cite{dai2018joint}.  We denote the local  energy consumption for computing task $D_i$ as  $E_{i}^l$, which can be defined as
\begin{equation}
\label{local_energy}
E_{i}^l={\varsigma c_i( {f_i^{l}})^2}.
\end{equation}

 \textit{2) Offloading to SBS}: Several devices may be associated with the same SBS and offload their tasks to the SBS simultaneously such that each one of them is only assigned to a part of the computation resource of the SBS. We denote  by $F_j^s$ the entire computation resource of SBS $j$ and by $f_{ij}^s$ the computation resource of SBS $j$ allocated to device $i$ for processing task $D_i$. The total amount of assigned computation resource cannot exceed the entire computation resource of SBS $j$, i.e., $\sum_{i\in\mathcal{U}}y_i^sf_{ij}^s\leq F_j^s$.

The task completion delay in this case consists of three parts. The first one is  the uplink wireless transmission delay  for offloading task $D_i$ from device $i$ to SBS $j$. The second one is the task computing delay which is related to the allocated computation resource $f_{ij}^s$. The third part is the downlink wireless transmission delay of the computation result from SBS $j$ to device $i$. Since the size of the result is often much smaller than the input data size, we ignore the duration of the downlink transmission. The task completion delay  for offloading task $D_i$ from device $i$ to SBS $j$ can be written as
\begin{equation}
\label{time_SBS}
T_{ij}^s={d_{i}}/{R_{ij}^s}+{c_i}/{f_{ij}^{s}},
\end{equation}

The total energy  consumption  for completing task $D_i$   is given by 
\begin{equation}
\label{energy_SBS}
E_{ij}^s={p_id_{i}}/{R_{ij}^s}+{c_i}*{e_j},
\end{equation}
where $e_j$ is the  energy consumption per  computation resource of SBS $j$.

 \textit{3) Offloading to the MBS}: If device $i$ chooses to offload its task to the MBS, the task execution process is similar to the  case of offloading to an SBS. Thus, the  task completion delay for offloading task $D_i$ from device $i$ to the MBS  can be expressed as
 \begin{equation}
 \label{time_MBS}
T_{i0}^m={d_{i}}/{R_{i0}^m}+{c_i}/{f_{i0}^{m}},
 \end{equation}where $f_{i0}^{m}$ is the computation resource of the MBS assigned to device $i$. The  total amount of assigned computation resource cannot exceed the entire computation resource of the MBS, i.e., $\sum_{i\in\mathcal{U}}z_i^mf_{i0}^m\leq F_0^m$, where $F_0^m$ denotes the entire  computation resource of the MBS. The corresponding energy  consumption  for completing  task $D_i$ is $E_{i0}^m={p_id_{i}}/{R_{i0}^m}+{c_i}*{e_0}$, where $e_0$ is the  energy consumption per  computation resource of the MBS.
 
 \section{Problem Formulation}
\label{pF}
The objective of  the joint computation offloading and resource allocation problem is to minimize system energy consumption. Taking the constraints of  computation resources into account, we formulate the  optimization problem as follows:

\begin{subequations}
	\label{S1}
	\begin{align}
	\centering
		\min	~~& \sum_{i\in\mathcal{U}}\sum_{j\in\mathcal{B}} x_i^lE_i^l+y_{ij}^sE_{ij}^s+z_i^mE_{i0}^m  \tag{8} \\    
			s.t.~
	          &\sum_{j\in\mathcal{B}} y_{ij}^sT_{ij}^s + x_i^lT_i^l+z_i^mT_{i0}^m	\leq  \tau_i, \label{c1}\\ 
			&~\sum_{j\in\mathcal{B}}y_{ij}^s+ x_i^l+z_i^m=1,\label{c2}\\    	
			&~\sum_{i\in\mathcal{U}}y_{ij}^sf_{ij}^s\leq F_j^s, ~\sum_{i\in\mathcal{U}}z_{i}^mf_{i0}^m\leq F_0^m, \label{c4}\\
			&~0\leqslant f_{ij}^s \leqslant F_j^s,0\leqslant f_{i0}^m \leqslant F_0^m,\label{c5}\\
			&~x_{i}^l,y_{ij}^s,z_i^m\in\{0,1\} \label{c7} 
	\end{align}
\end{subequations}

Constraint (\ref{c1}) guarantees that the task processing delay cannot exceed the upper limit  $\tau_i$. Constraints (\ref{c2})  and (\ref{c7}) together ensure that each device only chooses one way to compute its task,i.e., executing locally or offloading to an SBS or the MBS.  Constraint (\ref{c4}) ensures that the sum of the computation resources of each base station allocated to all tasks does not exceed the total amount of computation resources the base station has.  Because of the integer constraint (\ref{c7}), problem (\ref{S1}) is  NP-hard \cite{Garey:1990:CIG:574848}.


We attempt to exploit DRL to  design a strategy for minimizing the system energy consumption.  Since DRL is based on Markov Decision Process (MDP), we  first transform problem (\ref{S1}) as MDP form and then propose a DRL-based algorithm to solve it.

 A typical MDP is defined by a 4-tuple $\mathcal{M}=(\mathcal{S},\mathcal{A},\mathcal{P},\mathcal{R})$, where $\mathcal{S}$ is a finite set of states, $\mathcal{A}$ is a finite set of actions, $\mathcal{P}$ represents transition probability from state $s$ to state $s'$ after executing action $a$, and $\mathcal{R}$ is an immediate reward function for taking action $a$. 

 \textit{1) System State}:
At the beginning of each time slot, the MBS observes  system states of  heterogeneous wireless networks which includes all task offloading requests of devices, the available computation resource of the MBS and SBSs,  locations of SBSs and devices, and  wireless  data rates between devices and base stations. State $s_t\in\mathcal{S}$ at time slot $t$ can be defined as 
\begin{equation}
s_t =  \{\mathbf{d}(t), \mathbf{c}(t),\mathbf{\tau}(t), \mathbf{R}^s(t), \mathbf{R}^m(t), \mathbf{F}(t)\},
\end{equation}
where
\begin{itemize}
\item $\mathbf{d}(t) = [d_1(t),...,d_N(t)]$:  represents input data sizes of  computation-intensive tasks  at time slot $t$;
\item $\mathbf{c}(t) = [c_1(t),...,c_N(t)]$: represents the remaining required resources for completing the  computation-intensive tasks  at time slot $t$;
\item $\mathbf{\tau}(t) = [\tau_1(t),...,\tau_N(t)]$:  represents the maximal delay of each computation-intensive  tasks   at time slot $t$;

\item $ \mathbf{R}^s(t) =\{[R_{11}^s(t),...,R_{N1}^s(t)],...,[R_{NM}^s(t),...,R_{NM}^s(t)]\}$, represents wireless data rates between devices and SBSs  at time slot $t$;

\item $ \mathbf{R}^m(t) =\{ [R_{10}^m(t),...,R_{N0}^m(t)]\}$, represents  wireless data rates between devices and the MBS  at time slot $t$;

\item $ \mathbf{F}(t)= [F_0^m(t) ,F_1^s(t),..,F_M^s(t)]$: represents the available computation resource of  the MBS and SBSs at time slot $t$, respectively;


\end{itemize}

 \textit{2) System Action}:
The  joint computation offloading and resource allocation problem  consists of two main components:  \textit{offloading decision-making} and \textit{computation resource  allocation}.  Action $a_t\in\mathcal{A}$ in time slot $t$ is defined as 
\begin{equation}
a_t = \{\mathbf{x}(t),\mathbf{y}(t),\mathbf{z}(t),\mathbf{f^s}(t),\mathbf{f^m}(t)\},
\end{equation}
where 
\begin{itemize}
\item $\mathbf{x}(t) = [x_1^l(t),...,x_N^l(t)]$:  represents offloading decisions  whether to compute tasks locally, i.e., if $x_i^l(t)=1$, task $D_i$ will be executed locally at time slot $t$,
 
\item   $\mathbf{y}(t) = [y_{11}^s(t),...,y_{1M}^s(t),...y_{N1}^s(t),...,y_{NM}^s(t)]$: denotes offloading decision about whether or not to  offload tasks to SBSs, i.e., if  $y_{ij}^s(t) =1$, task $D_i$ will be offloaded to SBS $j$ t,
\item  $\mathbf{z}(t) = [z_1^m(t),...,z_N^m(t)]$:  means offloading decision about whether or not to  offload tasks to the MBS, i.e., if $z_i^m(t)=1$, task $D_i$ will be offloaded and executed at the MBS at time slot $t$,
\item $\mathbf{f^s}(t) =[f_{11}^s(t),...,f_{1M}^s(t),...f_{N1}^s(t),...,f_{NM}^s(t)]$:  represents the  computation resource that each SBS assigns to each device at time slot $t$. 
\item $\mathbf{f^m}(t) =[f_{10}^m(t),...,f_{N0}^m(t)]$:  denotes the  computation resource that the MBS assigns to each device at time slot $t$. 
\end{itemize}

 \textit{3) Reward Function}: The objective of joint computation offloading and resource allocation problem  is to minimize system energy consumption. After an action $a_t$ is taken, the system  computes  immediate reward $\mathcal{R}^{imm}(s_t,a_t)$ and then  updates system state.   Thus, the  immediate reward function can be defined as
\begin{equation}
\label{reward1}
\begin{split}
\mathcal{R}^{imm}(s_t,a_t )= -\mathbb{E}[\sum_{i\in\mathcal{U}}\sum_{j\in\mathcal{B}}& x_i^l(t)E_i^l(t)+y_{ij}^s(t)E_{ij}^s(t)\\&+z_i^m(t)E_{i0}^m(t)],
\end{split}
\end{equation}
where $E_i^l(t)$, $E_{ij}^s(t)$, and $E_{i0}^m(t)$ denote local energy consumption for executing task $D_i$,  energy consumption for processing task $D_i$ when offloaded to SBS $j$, and  energy consumption for computing task $D_i$  when offloaded to the MBS, respectively, at time slot t.   


 \textit{4) Next State}: After computing  immediate reward, the system will update state from $s_t$ to $s_{t+1}$.  As the system performs computation offloading and resource allocation in time slot $t$,  the state  that needs to be updated is  $\mathbf{d}(t), \mathbf{c}(t),\mathbf{\tau}(t), \mathbf{F}(t)$. 
   
   If device $i$ chooses to compute its task locally (i.e., $x_i^l =1$), $ c_i(t)$ and $\tau_i(t)$ need to be updated based on Eq. (\ref{local_time}), the computed task during time slot $t$ is $f_i^l*\bigtriangleup t$, where $\bigtriangleup t$ is the length of each time slot. $ c_i(t+1) = c_i(t) -f_i^l* t$ and $\tau_i(t+1) =\tau_i(t) -  t$. Since local computing does not involve wireless transmission, there is no time computation and energy computation on wireless channels such that we consider $d_i(t) = 0$ in time slot $t$.
   
   If device $i$ chooses to offload its task to SBS $j$ (i.e., $y_{ij}^s =1$),  $ d_i(t)$, $ c_i(t)$, and $\tau_i(t)$ are updated based on  Eq. (\ref{time_SBS}). Specifically, $ d_i(t+1) =d_i(t) -R_{ij}^s*t $, $ c_i(t+1)= c_i(t)-f_{ij}^s*t$, and  $\tau_i(t+1) =\tau_i(t) -  t$. The available  computation resource of  each SBS is also updated. Specifically, $F_{j}^s(t+1) = F_{j}^s(t)- \sum_{i\in \mathcal{U}}y_{ij}^s*f_{ij}^s$  \cite{8377343}. If device $i$ chooses to offload its task to the MBS, the updated process is similar to the case of offloading to an SBS.

  The objective of the joint computation offloading and resource allocation is to maximize the long-term revenue of the wireless networks, i.e., to maximize  the cumulative total reward, given by 
 \begin{equation}
 \label{long}
\mathcal{R} = \max \mathbb{E}\left[\sum_{t=0}^{T-1} \varepsilon^t\mathcal{R}(s_t,a_t )\right],
 \end{equation}
 where $\varepsilon\in[0,1]$ is the discount factor. If all tasks  can be completed  within their stringent delay constraints, the system agent gets a total reward $\mathcal{R}$. Otherwise, the agent receives a penalty,  which is a negative  constant. 
 \begin{figure}
 	\centering
 	\includegraphics[width =3.50 in]{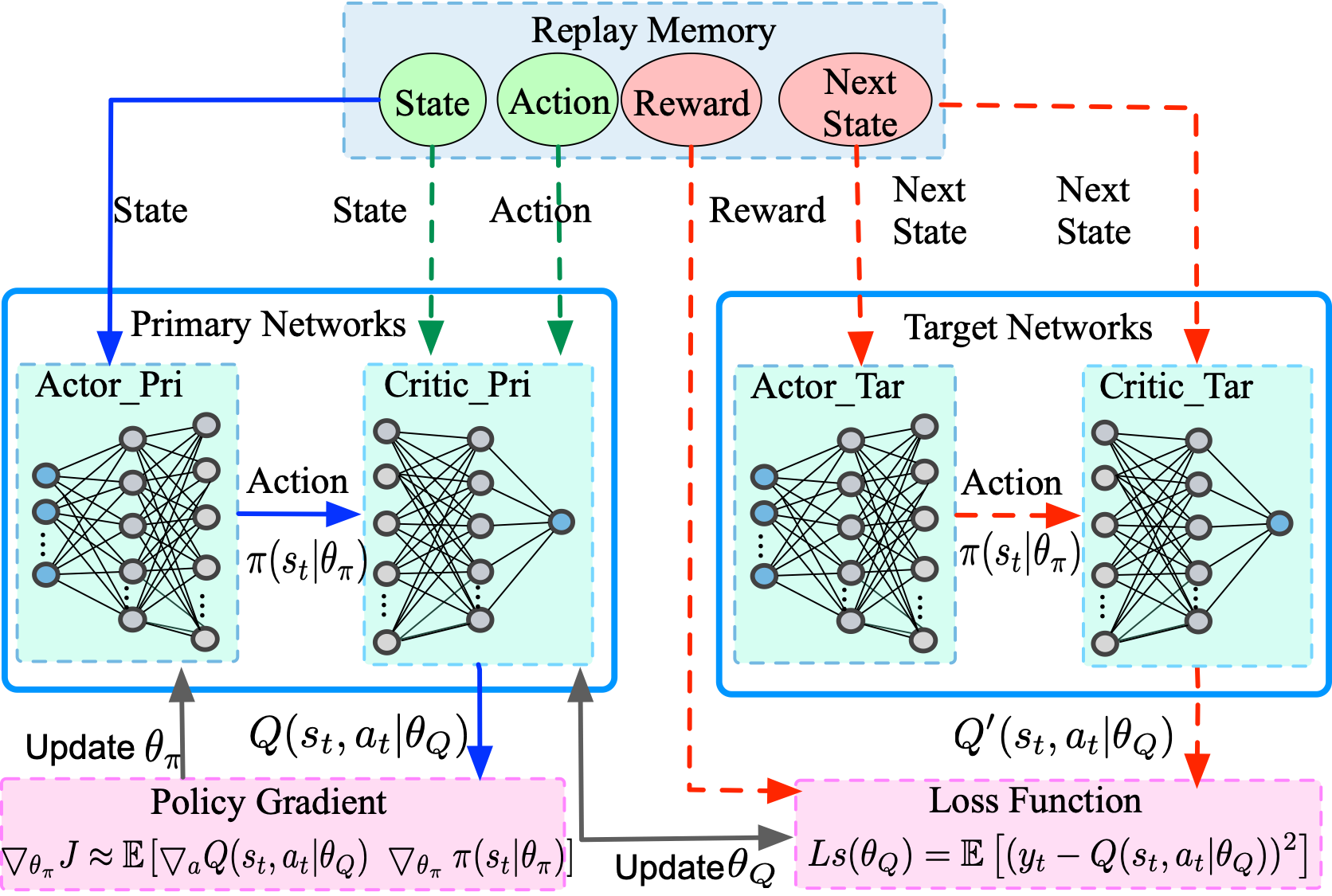}
 	\caption{The structure of DDPG}
 	\label{DDPG}
 \end{figure}

\section{ DRL-based Computation Offloading and Resource Allocation Scheme}

Since the action space of the proposed problem involves continuous variable, we utilize actor-critic based DRL algorithm, Deep Deterministic Policy Gradient (DDPG) to find the solution of  MDP and then  we  incorporate action refinement in DRL to make jointly offloading and resource allocation. In this section, we first introduce the DDPG-based computation offloading and resource allocation algorithm. Then, we describe the way of action refinement.

\label{DRL}
  \subsection{DDPG-based Computation Offloading and Resource Allocation Algorithm}
  AI  is a promising approach to facilitate in-depth feature  discovery such that  the dynamic input fields for edge computing and caching problems can be obtained in advance. In addition, AI is also a promising tool to tackle  complex  optimization problems, such as  resource allocation. DRL is a branch of AI research where an agent learns by interacting with  the environment.  Here, we exploit the advanced DRL algorithm-DDPG to solve (\ref{long}). DDPG is an \textit{Actor-Critic} framework based algorithm, where \textit{Actor} is used to generate actions and \textit{Critic} is used to guide the \text{actor}  to produce better actions.  The DDPG consists of three modules: primary network, target network, and replay memory, as shown in Fig. \ref{DDPG}.  Primary network  generates the detailed computation offloading and resource allocation  policy by mapping current state $s_t$ to an action $a_t$.  Primary network  consists of two Deep Neural Networks (DNN), namely primary actor DNN $\pi(s_t|\theta_{\pi})$ and  primary critic DNN $Q(s_t,a_t|\theta_{Q})$. Target network is used to generate  target values for training  primary critic DNN.  The structure of target network is the same as the structure of  primary network but with different parameters, the target network is denoted by  $\pi'(s_t|\theta_{\pi}^T)$ and $Q'(s_t,a_t|\theta_{Q}^T)$. Replay memory is used to store experience tuples. Experience tuples  include current state $s_t$, the selected action $a_t$, reward $\mathcal{R}^{imm}(s_t,a_t )$, and the next state $s_{t+1}$. The random experience fetched from the replay memory breaks up the correlation among the experiences in a mini-batch.


1)  \textit{Primary Actor DNN Training:} 

The  explored policy can be defined as a function parametrized  by  $\theta_{\pi}$, mapping current state to  an action $\hat{a} = \pi(s_t|\theta_\pi)$ where  $\hat{a}$ is a proto-actor action generated by the mapping and $\pi(s_t|\theta_\pi)$ is the explored edge caching and content delivery policy produced by DNN. By adding an Ornstein-Uhlenbeck noise $\mathfrak{N}_t$, the constructed action can be described as \cite{lillicrap2015continuous}
 \begin{equation}
 \label{action}
a_t = \pi(s_t|\theta_\pi)+\mathfrak{N}_t.
 \end{equation} 
{where Ornstein-Uhlenbeck noise $\mathfrak{N}_t$ means the random exploration of DDPG. }

 The primary actor DNN updates network parameter  $\theta_{\pi}$ using the sampled policy gradient  as,
  \begin{equation}
  \label{pg}
\bigtriangledown_{\theta_{\pi}}J\approx \mathbb{E} \left[\bigtriangledown_{a}Q(s,a|\theta_{Q})|_{s=s_t,a=\pi(s_t)}   \bigtriangledown_{\theta_{\pi}} \pi(s|\theta_\pi)|_{s=s_t}\right],
  \end{equation}
 where $Q(s,a|\theta_{Q})$ is an action-value function  and will be introduced in the following. Specifically, at each training step, $\theta_{\pi}$ is updated by a mini-batch experience $<s_t,a_t,\mathcal{R}^{imm},s_{t+1}>,$ $ t \in \{1,...,V\}$, randomly sampled from the replay memory,
 \begin{equation}
 \small
\theta_{\pi} = \theta_{\pi}-\frac{\alpha_{\pi}}{V}\sum_{t=1}^{V}\left[\bigtriangledown_{a}Q(s,a|\theta_{Q})|_{s=s_t,a=\pi(s_t)} \bigtriangledown_{\theta_{\pi}} \pi(s|\theta_\pi)|_{s=s_t}\right],
 \end{equation}
 where $\alpha_{\pi}$ is the learning rate of the primary actor DNN. 
 \begin{algorithm}[!tbp]
 
 	\caption{ DDPG-based Content Sharing algorithm}
 		\label{outer}
 	\begin{algorithmic}[1]
 	\State Initialize  the primary actor DNN $\pi(s|\theta_\pi)$ and the primary critic DNN  network $Q(s_t,a_t|\theta_{Q}) $ with parameters $\theta_\pi$ and $\theta_Q$;
 	\State Initialize the target actor DNN $\pi'(s|\theta_\pi^T)$ and the target critic DNN  network  $Q'(s,a|\theta_Q^T)$ with parameters $\theta_{\pi}^T\leftarrow\theta_{\pi}$ and $\theta_{Q}^T\leftarrow\theta_{Q}$;  Initialize replay memory;	        
     \For {each episode}
 		\State Initialize  environment setup;
 	 	\For {each time slot $t$ }
 	     	\State Execute  action $a_t$ based on   (\ref{action}).
 	        \State Observe reward $\mathcal{R}^{imm}(s_t,a_t)$ based on   (\ref{reward1}) and  update  state $s_{t+1}$;
 	        \State Store  tuple $<s_t,a_t,\mathcal{R}^{imm}(s_t,a_t),s_{t+1}>$ into replay memory;
 	        \State Sample a mini-batch of  $V$ tuples from replay memory;
 	        \State Compute target value $y_t$ based on  (\ref{target});
 	        \State Update the primary critic DNN  network parameter $\theta_Q$  by minimizing  loss function (\ref{loss});
 	        \State Update  the primary actor DNN network parameter $\theta_{\pi}$ using the sampled policy gradient (\ref{pg});
 	        \State Update the target network parameters:
 	        \NoNumber{~~~~~~~~$\theta_{\pi}^T\leftarrow\omega\theta_{\pi}+(1-\omega)\theta_{\pi}^T$}
 	     \NoNumber{~~~~~~~~$\theta_{Q}^T\leftarrow\omega\theta_{Q}+(1-\omega)\theta_{Q}^T$}         	         
 		\EndFor 
     \EndFor
 	\end{algorithmic}
 \end{algorithm}

2)  \textit{Primary Critic DNN Training:}  

The primary critic DNN evaluates the performance of the selected action  based on the action-value function.  The action-value function is calculated by the Bellman optimality equation and can be expressed as 
 \begin{equation}
Q(s_t,a_t|\theta_{Q}) = \mathbb{E} \left[\mathcal{R}^{imm}(s_t,a_t)+\varepsilon Q(s_{t+1},\pi(s_{t+1})|\theta_{Q}) \right],
 \end{equation}
 Here, the primary critic DNN takes both  current state $s_t$ and next state $s_{t+1}$ as input to calculate $Q(s_t,a_t|\theta_{Q})$  for each action.

The primary critic DNN updates the network parameter  $\theta_{Q}$ by minimizing the loss function $Ls(\theta_{Q})$. The   loss function  is defined as 
 \begin{equation}
 \label{loss}
Ls(\theta_{Q}) = \mathbb{E} \left[(y_t-Q(s_t,a_t|\theta_{Q}) )^2\right],
 \end{equation}
 where $y_t$ is the target value and can be obtained by
 \begin{equation}
 \label{target}
y_t = \mathcal{R}^{imm}(s_t,a_t)+\varepsilon Q'(s_{t+1},\pi'(s_{t+1}|\theta_{\pi}^T)|\theta_{Q}^T).
 \end{equation}
 where $Q'(s_{t+1},\pi'(s_{t+1}|\theta_{\pi}^T)|\theta_{Q}^T) $ is obtained through the target network, i.e., the network with parameters $\theta_{\pi}^T$ and $\theta_{Q}^T$. 
 
 The gradient of $Ls(\theta_{Q})$ is calculated as 
  \begin{equation}
  \label{loss_gra}
 \bigtriangledown_{\theta_{Q}}Ls = \mathbb{E} \left[2(y_t-Q(s_t,a_t|\theta_{Q}) )\bigtriangledown_{\theta_{Q}}Q(s_t,a_t)\right]
  \end{equation}
 At each training step, $\theta_{Q}$ is updated with a mini-batch experiences $<s_t,a_t,\mathcal{R}^{imm},s_{t+1}>,$ $ t \in \{1,...,V\}$, that randomly sampled from the replay memory,
  \begin{equation}
 \theta_{Q} = \theta_{Q}-\frac{\alpha_{Q}}{V}\sum_{t=1}^{V}\left[2(y_t-Q(s_t,a_t|\theta_{Q}) )\bigtriangledown_{\theta_{Q}}Q(s_t,a_t)\right],
  \end{equation}
  where $\alpha_{Q}$ is the learning rate of the primary critic DNN.

3)  \textit{Target Network Training:} 

The target network can be regarded as an old version of the primary network with different parameters $\theta_{\pi}^T$ and $\theta_{Q}^T$. At each iteration, the parameters $\theta_{\pi}^T$ and $\theta_{Q}^T$ are updated based on the following definition:
  \begin{equation}
 \label{target_update}
\begin{split}
&\theta_{\pi}^T =\omega\theta_{\pi}+(1-\omega)\theta_{\pi}^T, \\
&\theta_{Q}^T =\omega\theta_{Q}+(1-\omega)\theta_{Q}^T ,
\end{split}
  \end{equation}
  where $\omega\in [0,1]$.

  The proposed DDPG-based computation offloading and resource allocation  algorithm is summarized as  Algorithm \ref{outer}. The algorithm first initializes the computation offloading and resource allocation  policy $\pi(s|\theta_\pi)$ of the primary actor DNN  with parameter $\theta_\pi$, and   initializes the action-value faction  $Q(s_t,a_t|\theta_{Q}) $ of the   primary critic DNN  network with parameter $\theta_Q$. Both parameters $\theta_{\pi}^T$ and $\theta_{Q}^T$ of the target network are also initialized. Then, according to current policy $\pi(s|\theta_\pi)$ and  state $s_t$, the primary actor DNN   generates action $a_t$   based on  (\ref{action}). According to the observed reward $\mathcal{R}^{imm}(s_t,a_t)$ and next state $s_{t+1}$,  a tuple $<s_t,a_t,\mathcal{R}^{imm}(s_t,a_t),s_{t+1}>$ is constructed and stored  into replay memory. Note that, if replay memory is going to be full, the oldest experience will be deleted to make room for the latest one. Based on mini-batch technique,  the algorithm  updates the primary critic DNN  network by minimizing  the function $Ls(\theta_{Q}) $  and   updates the primary actor DNN using the sampled policy gradient. After  a period of training, the parameters of the target networks are updated based on  Eq. (\ref{target_update}). 

   {The complexity of  Algorithm \ref{outer} is mainly determined by four neural networks and one activation layer.   Assuming that Actor DNN contains $L$ fully connected layers and Critic DNN contains $K$  fully connected layers.The time complexity can be calculated as \cite{qiu2019deep}
  
 \begin{equation}
 	\begin{split}
 	\centering
&  2\times \sum_{l=0}^{L} n_{Actor,l}\cdot n_{Actor,l+1} +2\times \sum_{k=0}^{K} n_{Critic,k}\cdot n_{Critic,k+1}\\
  & = \mathcal{O}(  \sum_{l=0}^{L} n_{Actor,l}\cdot n_{Actor,l+1} +\sum_{k=0}^{K} n_{Critic,k}\cdot n_{Critic,k+1})
  \end{split}
 \end{equation}
  where $n_{Actor,l}$ and $n_{Critic,k}$ mean the unit number in the $l$-th Actor DNN layer and the $k$-th Critic DNN layer, $n_{Actor,0}$ and $n_{Critic,0}$  equal the input size.}
  
 \subsection{Action Refinement}

The outputs of DDPG-based algorithm are continuous values. However, the actions about offloading decisions should be  integer values (i.e., $x_i^l, y_{ij}^s$ and $z_i^m$ ). Therefore,  { we adopt a rounding technique to refine these actions.} The rounding technique  consists of the following three steps: 1) obtain the continuous solution from $a_t$, 2) construct a weighted bipartite graph to establish the relationship between devices and base stations, 3) find an integer matching to obtain the integer solution.

1)  \textit {Offloading decisions abstraction:} We define $\mathbf{w} (t) \triangleq[\mathbf{x}(t),\mathbf{y}(t),\mathbf{z}(t)] $,  { thus each element $w_{ij}(t)$ in  $\mathbf{w} (t)$ is a continuous value.}

2)  \textit {Bipartite graph construction:}  {We  apply rounding technique \cite{shmoys1993approximation} to transform continuous value into integers by constructing the weighted bipartite graph $\mathcal{G}(\mathcal{U},\mathcal{V},\mathcal{E})$  to establish the relationship between devices and their offloading strategies.  $\mathcal{U}$ is one side of bipartite graph which represents devices in the network.
   $\mathcal{V} = \{v_{js}:j = 0,1,..,M, M+1, s= 1,...,J_j\}$ is the other side of  bipartite graph which is a set of virtual nodes. The index $j$ of $v_{js}$ is related to offloading strategies, where $j = 0$ means local computing, $j =\{ 1,...,M\}$  means  offloading to SBS, and $j=M+1$ means offloading to the MBS. The index of $s$ ranges from 1 to $J_j$, where $J_j = \lceil\sum_{i =1}^{N}w_{ij}(t)\rceil$) means there are $J_j$ devices will choose the $j$-th offloading  strategy. }

   The most important procedure for constructing graph $\mathcal{G}$ is  to set the edges and the edge weight between $\mathcal{U}$ and $\mathcal{V}$.  {The edges in $\mathcal{G}$ are constructed using  Algorithm 2. Specifically, if $J_j\leq 1$, there is only one node $v_{j1}$ corresponding to base station $j$. In this case, for each $w_{ij}(t)>0$, we add an edge between node $u_{i}$ and $v_{j1}$,  and set the weight of this edge as $w_{ij}(t)$. Otherwise, for each $s \in \{1,2,3...,J_j-1\}$, we  need to find the minimum index $i_s$ which satisfies $\sum_{i=1}^{i_s}w_{ij}(t)\geqslant s$. If $s = 1$, we set  $ i_{s-1} = 0$.  For each $i \in \{ i_{s-1}+1,..,i_{s}-1\}$ and $w_{ij}(t)>0$, we add  $(u_i,v_{js})$ into  $\mathcal{E}$  and set the weight of this edge as $w_{ij}(t)$. If $i = i_s$, we  add edge $(u_i,v_{js})$ into  $\mathcal{E}$ and set  the weight  as $s-$$\sum_{i=1}^{i_s-1}w_{ij}(t)$. This ensures that the sum of weights for all edges  $(u_i,v_{js})$ is equal to $s$.   If  $\sum_{i=1}^{i_s}w_{ij}(t)> s$, we add one more edge  $(u_i,v_{j(s+1)})$ and set the weight of this edge as $ \sum_{i=1}^{i_s}w_{ij}(t)-s$.}
  \begin{algorithm}[!tbp]
     	\caption{ Construct the edges of  bipartite graph $\mathcal{G}$}
     	\label{edge}
     	\begin{algorithmic}[1] 
     	\State Set $\mathcal{E}\leftarrow\varnothing$.
     	\If{$J_j\leqslant 1$ }
     	        \State There is only one node $v_{j1}$ corresponding to base 
     	        \NoNumber{station $j$. }
               	\For{ each $w_{ij}(t)>0$}
                     	\State Add edge $(u_i,v_{j1})$ into $\mathcal{E}$ and set { $e_{ij1} =  w_{ij}(t)$. }
             	\EndFor
     	\Else
     	       	\For{ each $s \in \{1,2,..,J_j-1\}$}
     	        \State Find the minimum index $i_s$ where  $\sum_{i=1}^{i_s}w_{ij}(t)\geqslant s$.
     	        \If {$s = 1$}
     	        \State $i_{s-1} = 0$
     	         \EndIf
     	        \For{$i \in \{i_{s-1}+1,..,i_{s}-1\}$ and $w_{ij}(t)>0$}
     	               \State Add  edge $(u_i,v_{js})$ into  $\mathcal{E}$ with $e_{ijs}  = w_{ij}(t)$.
     	        \EndFor
     	        \If{ $i = i_s$}
     	             \State Add edge $(u_i,v_{js})$ into  $\mathcal{E}$ with  $e_{ijs}  = s-$$\sum_{i=1}^{i_s-1}w_{ij}(t)$. 
     	        \EndIf
     	        \If{$\sum_{i=1}^{i_s}w_{ij}(t)> s$.}
     	              \State Add edge $(u_i,v_{j(s+1)})$ into  $\mathcal{E}$ with weight 
     	              \NoNumber{$e_{ij(s+1)} =  \sum_{i=1}^{i_s}w_{ij}(t)-s$.  }
     	      	\EndIf
     	     	\EndFor
     	\EndIf
     	\end{algorithmic}  
     \end{algorithm}

   
  3)  \textit {Action refinement:}  We utilize the Hungarian algorithm \cite{kuhn1955hungarian} to find a complete max-weighted 
  bipartite  matching $M_{match}$.  According to the  $M_{match}$, we obtain the integer offloading decisions. Specifically, if $(u_i,v_{js}, e_{ijs})$ is in the $M_{match}$, we set $w_{ij}(t)= 1$; otherwise, $w_{ij}(t)= 0$. Based on the definition $\mathbf{w} (t) = [\mathbf{x}(t),\mathbf{y}(t),\mathbf{z}(t)] $, we can decide the values of   $x_i^l(t)$, $y_{ij}^s(t)$. Thus, we get the binary offloading policy. 
   {The complexity of action refinement algorithm is polynomial in the number of nodes and edges,  that is $\mathcal{O(|\mathcal{V}||\mathcal{E}|)}$. }

\section{Numerical Results}
\label{sr}
In this section, we use Python and TensorFlow to evaluate the performance of our proposed DDPG-inspired computation offloading and resource allocation  algorithm based on a real-word dataset. 

\subsection{Simulation Setup}
 We consider a network topology with one MBS , $M=10$   SBSs, and $N=100$ devices.  The maximum transmission power of devices is set to $100$ mW.  The channel gain models presented in 3GPP standardization are adopted here \cite{ikuno20103gpp}. The noise power is $\sigma^2 = 10^{-11}$ mW. The bandwidth of the MBS and the SBSs are $10$ MHz  and $5$ MHz, respectively.  In addition, $e_j=e_0 = 1$ W/GHz \cite{wittmann2016chip}. The CPU computation capabilities of devices, SBS, and MBS are $ 0.5$, $10$, and $50$ GHz, respectively. Each device has a computation task. 
   Both the data size of each task and required number of CPU cycles per bit follow the uniform distribution with $ U [5, 50]$ MB and $ U [0.5, 5]$ GHz, respectively.
  The upper limit on latency is set as $1$ s. 
  
 Our proposed DDPG-inspired algorithm to the above pattern recognition tasks  is deployed  on a MacBook Pro laptop, powered by two Intel Core i5 processor (clocked at 2.6GHz).  The  activation function of the DDPG  is $\frac{tanh(x)+1}{2}$. The size of mini-batch is set as 32. The maximum number of episodes is 6000 and the maximum number of steps in each episode is set to 20.  The penalty is $100*N$, where $N$ is the number of devices. 
 
 To verify the performance of our proposed algorithm, we introduce the following two benchmark policies:
 \begin{itemize}
 \item \textbf{Local  computing:}  Each mobile user  executes its computation tasks locally. The  feasible computation resource of each task can be obtained as $f_i^k = \frac{F_id_i^kc_i^k}{\sum_{k=1}^{K}d_i^kc_i^k}$. If  $\sum_{k=1}^{K}\frac{d_i^kc_i^k}{f_i^k} \leqslant T_i^{max}$, 
 all tasks are always locally executed.
 
 \item \textbf{Full offloading}: All computation  tasks are offloaded to the MBS for remote computing. In this policy, all mobile users are associated with the MBS. 

\begin{table}[!hbt]
	\caption{Different Types of Tasks }
	\label{table_2}	
	\centering
	\begin{tabular}{|l|l|l|l|}
		\hline
	&	\bfseries  Data size& \bfseries Required CPU cycles & \bfseries  Types\\
		\hline	   
		   
		Type 1	&50 MB &5 GHz& large data size \\
	             	&          &         & computation-intensive\\
		\hline
	   Type 2	&50 MB &0.5 GHz& large data size\\
	   	&          &         & non-computation-intensive\\
	  		\hline		
	  	 Type 3	&5 MB &5 GHz&  small data size\\
	  	 	&          &         & non-computation-intensive\\
	  		  		\hline																			
	\end{tabular} 
\end{table}

\end{itemize}
   \begin{figure}
   	\centering
   	\includegraphics[width =3.5 in]{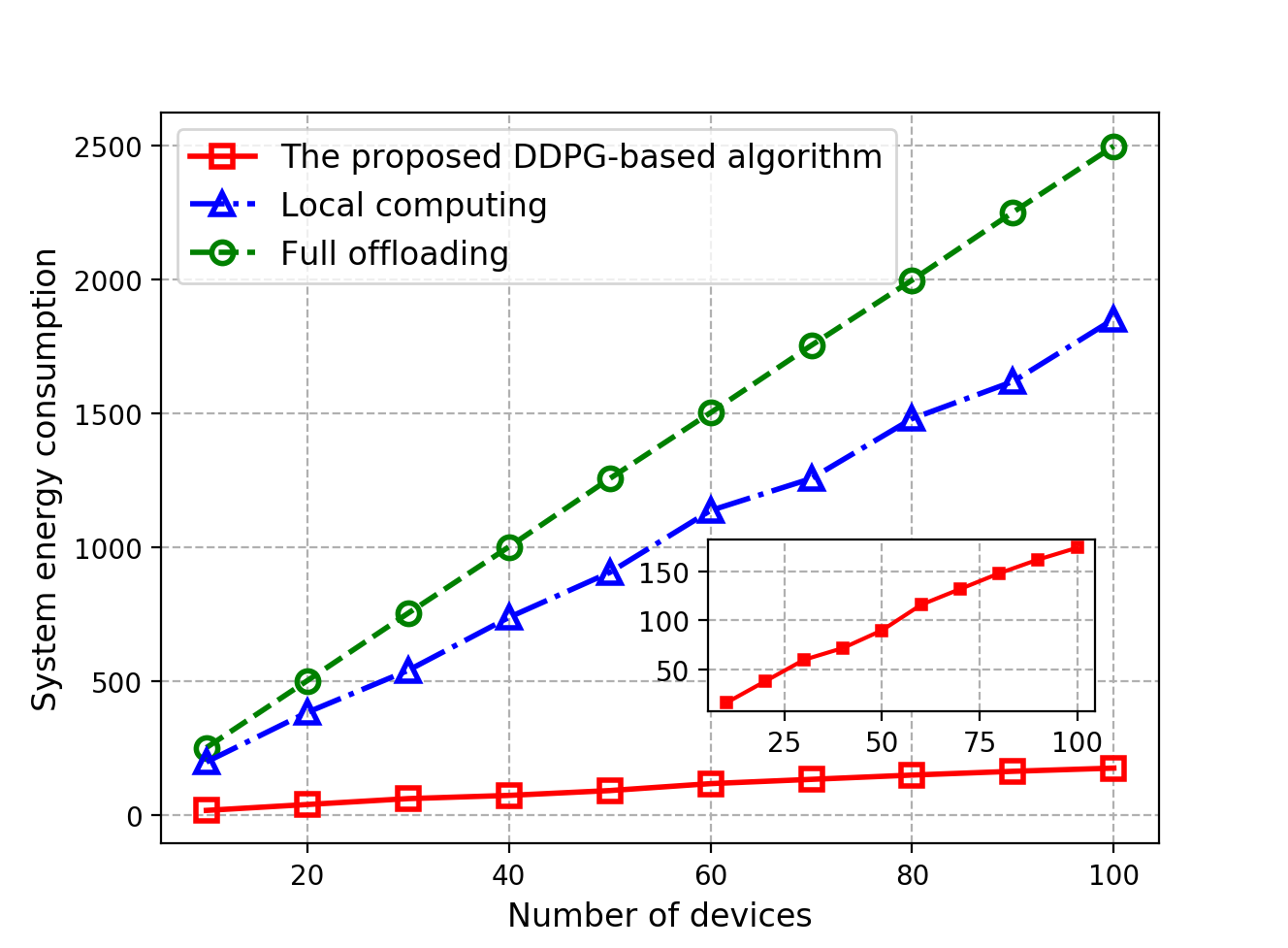}
   	\caption{Comparison of system energy consumption with respect to the number of devices under different schemes.}
   	\label{cfig1}
   \end{figure} 
   
\subsection{Performance Analysis}

\begin{figure}[!t]
         	\centering
         	\includegraphics[width =3.5 in]{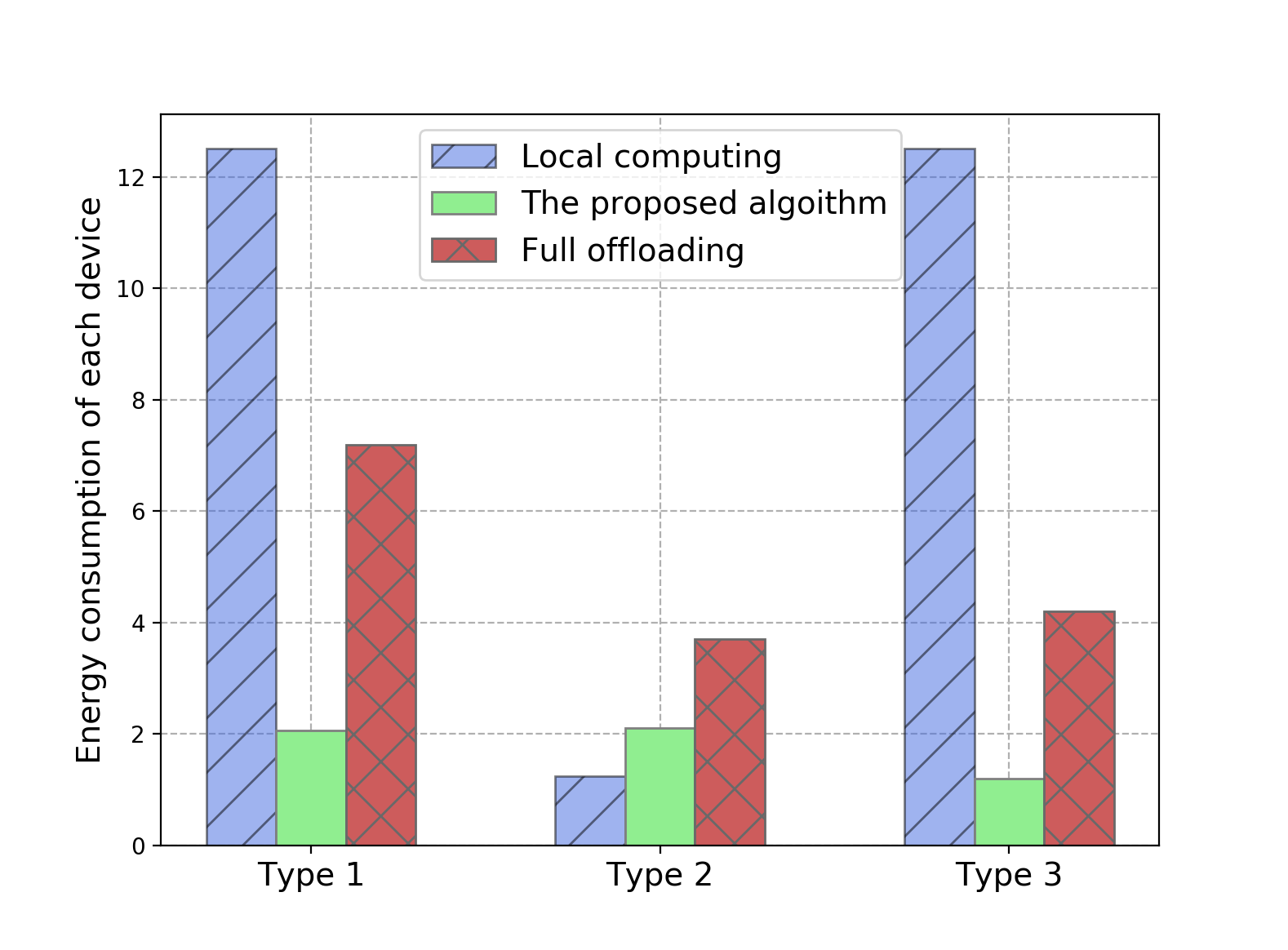}
         	\caption{Comparison of  energy consumption on each device  with respect to different task types.}
         	\label{cfig2}
         \end{figure}

In this subsection, we compare the proposed DDPG-based algorithm with other two benchmarks. In Fig. \ref{cfig1}, we plot the energy consumption with respect to the number of devices under different schemes. 
To evaluate the performance of  different algorithms, we consider three types of tasks, as shown in Table \ref{table_2}.  Fig. \ref{cfig2} shows the  energy consumption of each device  with respect to different task types under different algorithms.

From Fig. \ref{cfig1},  we can draw several observations. First, the proposed  DDPG-based algorithm significantly outperforms the two benchmark policies by jointly optimizing offloading decisions and computation resources.  Specifically, compared to the local computing policy,   the proposed  algorithm can offload tasks to edge servers with a relatively low computation energy consumption. Compared to the full offloading policy,  the proposed algorithm ensures offloading each task to the  nearby edge node which results in  lower energy consumption for wireless transmission.  Second, the system energy consumption of local computing increases rapidly as the number of devices becomes large.  Third, full offloading policy performs the worst. Since  all devices offload their tasks to the MBS.  Full offloading policy results in a congested wireless uplink and a higher wireless transmission energy consumption. 

From Fig. \ref{cfig2}, we can see that when tasks are computation-intensive (i.e., type 1 and type 3), the energy consumption of the proposed algorithm is the lowest.  The reason is that utilizing the proposed algorithm,  computation-intensive tasks can intelligently be offloaded to nearby edge servers. Compared to the scheme of local computing, the energy consumption on each device is dramatic reduced.  Compared to the scheme of full offloading, the proposed algorithm can support more offloading action such that the energy consumption on each device can be decreased.  When tasks are type 2 (i.e., large data size but less computation resources required),  local computing policy results in the lowest energy consumption. This is because transmitting large data incurs higher energy consumption.  However, for local computing, the only energy consumption is due to computation. In this case, our proposed algorithm leads to a slightly higher energy consumption than local computing, but it performs much better than full offloading.  {Overall, our proposed algorithm keeps a relatively low energy consumption  for all types of tasks.} Thus, we can conclude that the proposed algorithm  is the most robust one, considering that a real  network will have mixed traffic.

 \begin{figure}[!t]
         	\centering
         	\includegraphics[width =3.5 in]{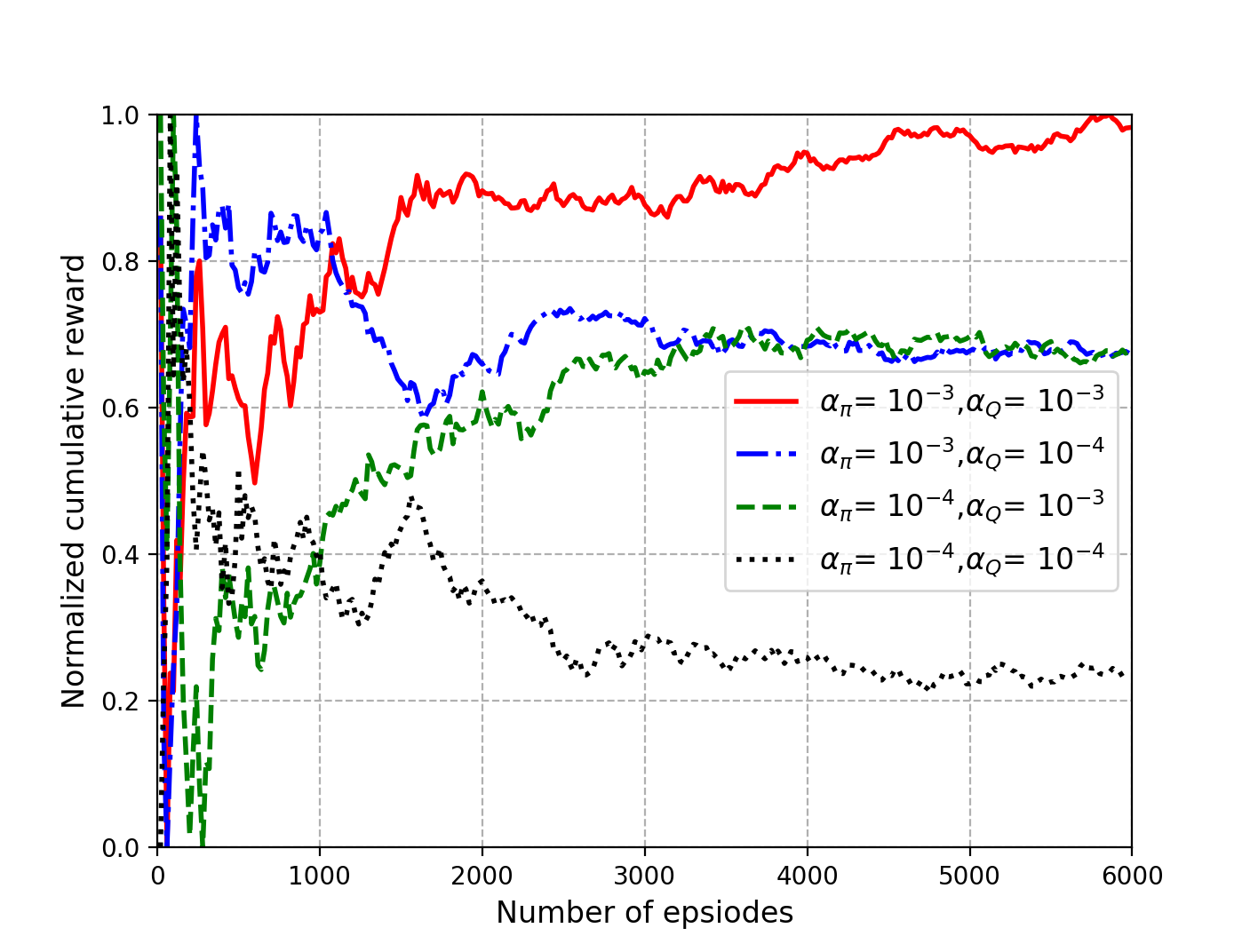}
         	\caption{Impact of learning rate}
         	\label{fig:64Kbpsa}
         \end{figure}

   \begin{figure}[!t]
           	\centering
           	\includegraphics[width =3.5 in]{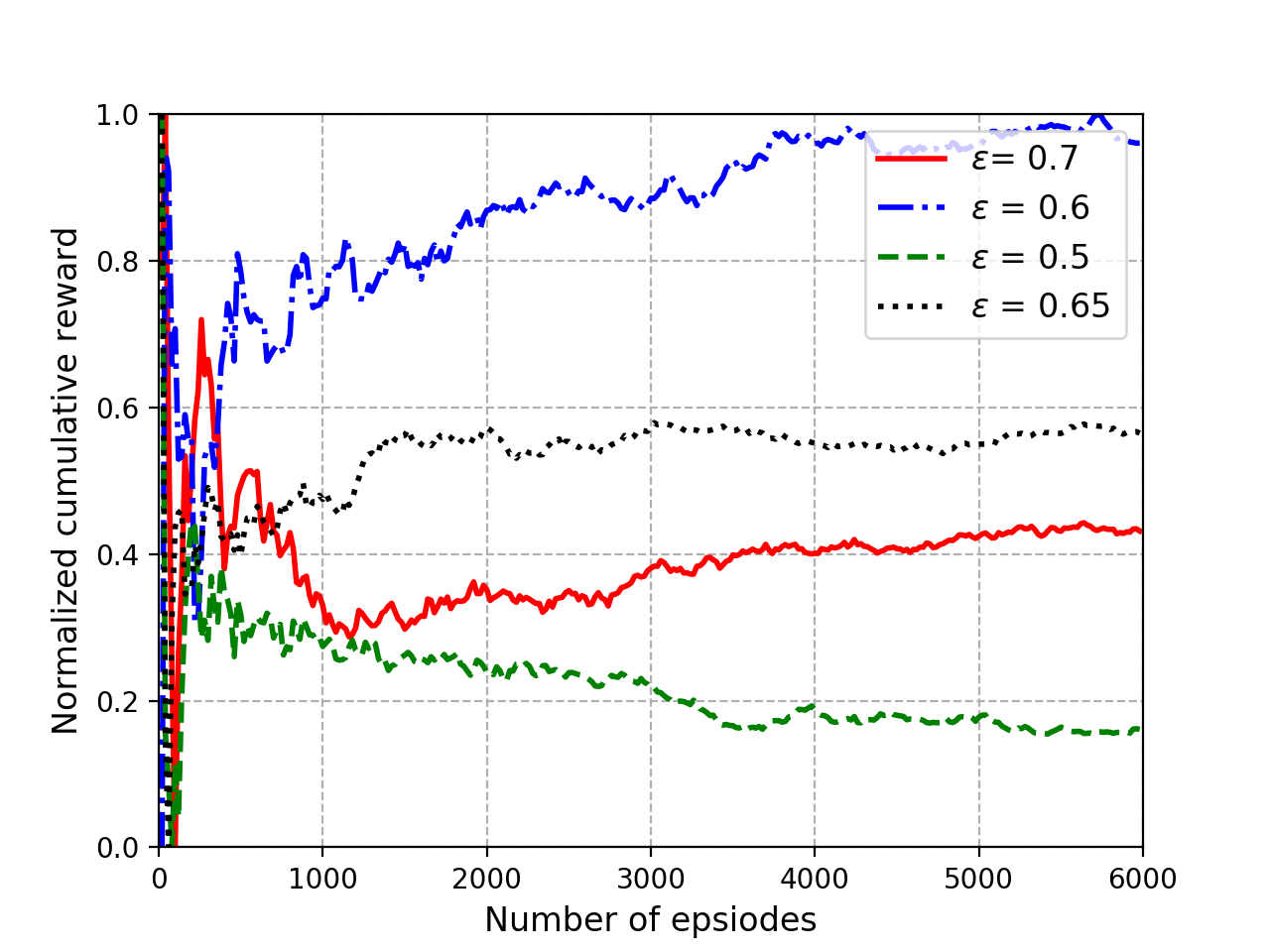}
           	\caption{Impact of discount factor}
           	\label{fig:64Kbpsb}
           \end{figure}

   \begin{figure}[!t]
           	\centering
           	\includegraphics[width =3.5 in]{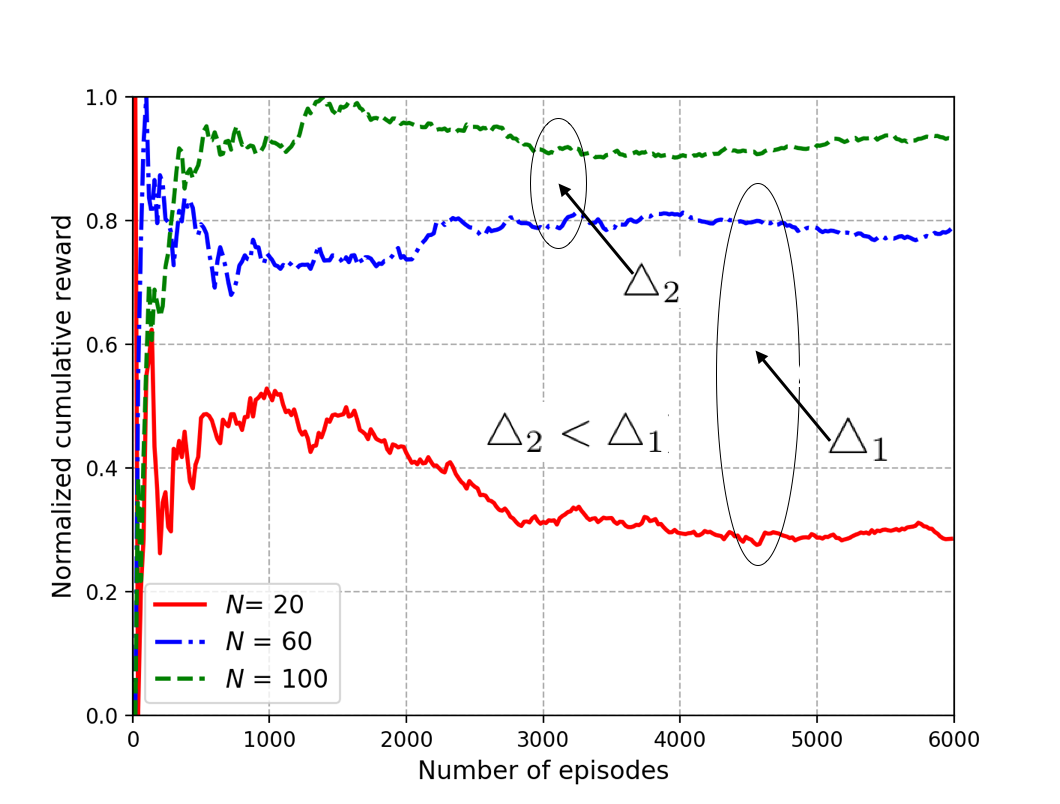}
           	\caption{Impact of number of devices}
           	\label{fig:64Kbpsc}
           \end{figure}

 \subsection{Impact of  Factors on Performance}  
 In this subsection, we evaluate the impact of  different factors on the performance of the proposed DDPG-based algorithm. The factors are: 1)  learning rates of  actor  DNN and critic DNN, 2) discount factor $\varepsilon$, and 3) number of devices. For ease of description, we adopt normalized cumulative reward in the y-axis. Here, a large normalized cumulative reward means a better performance. 
 
To  evaluate the impact of the learning rates, we set both $\alpha_{\pi}$ and $\alpha_{Q} $ as $\{10^{-2},10^{-3},10^{-4},10^{-5}\}$, respectively.  Fig. \ref{fig:64Kbpsa} depicts the normalized cumulative reward of the proposed DDPG-based algorithm under  learning rates.
Since only when $\alpha_{\pi}$ and $\alpha_{Q} $  are equal to $10^{-3}$ or $10^{-4}$ the proposed DDPG-based algorithm  achieves convergence, we only show the results of these cases in  Fig. \ref{fig:64Kbpsa}. We can draw several observations  from Fig. \ref{fig:64Kbpsa}.  {First, when $\alpha_{\pi} =10^ {-3} $,  the reward of red line (i.e., $\alpha_{Q} =10^ {-3} $) is greater than the reward of  blue line (i.e., $\alpha_{Q} =10^ {-4} $). When $\alpha_{\pi} =10^ {-4}$, the reward of green line (i.e., $\alpha_{Q} =10^ {-3} $) is greater than the reward of  black line (i.e., $\alpha_{Q} =10^ {-4} $). Similar,   when $\alpha_{Q} =10^ {-3} $,  the reward of red line (i.e., $\alpha_{\pi} =10^ {-3} $) is greater than the reward of  green line (i.e., $\alpha_{\pi} =10^ {-4} $). When $\alpha_{Q} =10^ {-4}$, the reward of blue line (i.e., $\alpha_{\pi} =10^ {-3} $) is greater than the reward of  black line (i.e., $\alpha_{\pi} =10^ {-4} $). Thus,  when  $\alpha_{\pi} =10^ {-3} $ and  $\alpha_{Q} =10^ {-3} $, the performance of the proposed algorithm is the best.} Second, when $\alpha_{\pi} = 10^ {-3} $ and $\alpha_{Q} = 10^ {-4} $, and $\alpha_{\pi} = 10^ {-4} $ and $\alpha_{Q} =10^ {-3} $,  the proposed algorithm eventually converges to the same value.  That means  $\alpha_{\pi}$ and $\alpha_{Q} $ have the same influence on the convergence. Third, when $\alpha_{Q} = 10^ {-3} $, the convergence trend of the red solid line  is similar to that of the green dash line. When $\alpha_{Q} = 10^ {-4} $  the convergence trend of  the blue dash-dot line and the black dot line is similar. Thus, the critic learning rate $\alpha_{Q}$ considerably influences the convergence trend. 

To  evaluate the impact of  discount factors,  we set $\varepsilon$ from $0.5$ to $0.7$.  Fig. \ref{fig:64Kbpsb} plots  the normalized cumulative reward of the proposed algorithm under different  discount factors. First, the proposed algorithm converges irrespective of the values of discount factor chosen. Second, when $\varepsilon = 0.6$, the normalized cumulative reward  is clearly higher than the cases when $\varepsilon = 0.65$ and $\varepsilon = 0.7$, which implies that a small discount factor results in a better performance. However, the performance when $\varepsilon = 0.65$ is also better than that when $\varepsilon = 0.5$ . Thus, we can conclude that $\varepsilon = 0.6$ is the best  discount factor for the proposed  algorithm. In fact, since $\mathcal{R} = \sum_{t=0}^{T-1} \varepsilon^t\mathcal{R}(s_t,a_t )$, the discount factor determines the relative ratio of  future immediate reward versus current immediate reward. Specifically, rewards received at the $t$-th time slot in the future are discounted exponentially by a factor of $\varepsilon^t$. Note that if we set $\varepsilon =0 $ , only current immediate reward  is considered. As we set  $\varepsilon $ closer to 1, future  immediate rewards are given greater weight relative to  the current immediate reward.

Fig. \ref{fig:64Kbpsc} shows the impact of the number of devices on the normalized cumulative reward. Here, we set $N$ as $\{20, 60,100\}$, respectively. We can see that the proposed algorithm converges  in all cases and the normalized cumulative reward increases with the increase in $N$. The growth in the number of devices  leads to more offloading requests, which results in the need and consumption of more communication resource and computation resource. Moreover, the gap between the normalized cumulative reward when $N=100$ and the normalized cumulative reward when $N=60$ is greater than the gap  between the normalized cumulative reward when $N=60$ and the normalized cumulative reward when $N=20$ (i.e., $\bigtriangleup_2<\bigtriangleup_1$), which means the increment of the normalized cumulative rewards becomes smaller. This observation implies that  the gain offered by the proposed algorithm in terms of reducing system energy consumption, is more pronounced when the number of devices is high. 


\begin{table}[!htbp]
\centering
	\caption{Training Delay of  the Proposed Algorithm }
	\label{table_3}	
\begin{tabular}{|c|c|c|c|}
\hline
\diagbox{Episodes}{Delay(s)}{$N$}&$20$&$40$&$60$\\ 
\hline
$3000$&35.17&38.31&51.40\\
\hline
$6000$&71.35&77.87&100.18\\
\hline
\end{tabular}
\end{table}

{Table \ref{table_3}	shows the training delay of the proposed algorithm, where $N$ is the number of devices. The number of devices determines the scale of state space and action space. Thus, as the number of devices increases, the training delay  increases dramatically. Further, with the increase in the number of episodes,  the training delay also increases.


       \begin{figure}[!h]
           	\centering
           	\includegraphics[width =3.3 in]{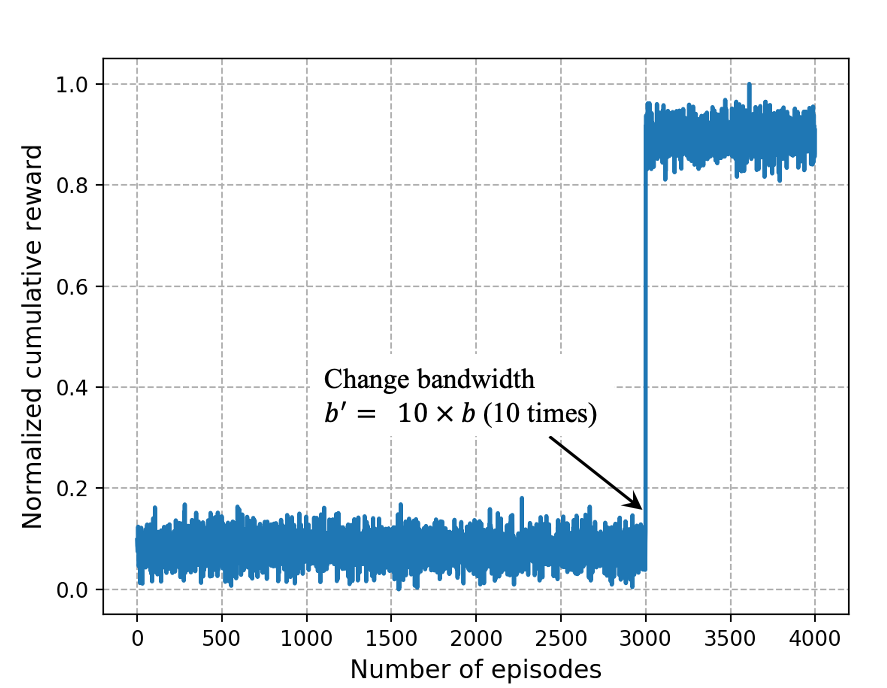}
           	\caption{Convergence under time-varying bandwidth}
           	\label{fig:6}
           \end{figure}

In Fig. \ref{fig:6}, we study the effect of time-varying parameter on the convergence of the proposed algorithm. To quickly verify the effect of  time-varying parameters on the proposed algorithm,  we consider a small-scale network with 10 devices and 5 SBSs. The bandwidth of the MBS and the SBSs are $2$ MHz  and $1$ MHz, respectively.  After running 3000 episodes, we change the bandwidth of base stations. Specifically,  we enlarge the original bandwidth ten times. As shown in Fig. 8, the normalized cumulative reward sharply increases  after the bandwidth is changed, and stabilizes at a higher value. This means the proposed algorithm automatically updates its offloading policy and converges to a new optimal solution.  Thus,  the proposed algorithm can deal with time-varying network environment.
}

\section{Conclusions}
\label{c}
In this paper, we proposed a joint  computation offloading and resource allocation scheme for minimizing system energy consumption in  5G heterogeneous networks. We first presented a multi-user   end-edge-cloud  orchestrated network in which  all devices and base stations have computation capabilities. Then, we formulated the joint  computation offloading and resource allocation  problem as an optimization problem and transformed it into the form of an MDP. We proposed a DDPG-based algorithm   to intelligently minimize system energy consumption by interacting with the environment. Numerical results based on a real-world dataset  demonstrated that the proposed  DDPG-based algorithm  significantly outperforms the benchmark policies  in terms of system energy consumption. Extensive simulations indicated that the  learning rate, discount factor, and number of end user devices have considerable  influence on  the performance of the proposed  DDPG-based algorithm.

	\bibliographystyle{IEEEtran}
	\bibliography{reference}

\begin{thebibliography}{10}
\providecommand{\url}[1]{#1}
\csname url@samestyle\endcsname
\providecommand{\newblock}{\relax}
\providecommand{\bibinfo}[2]{#2}
\providecommand{\BIBentrySTDinterwordspacing}{\spaceskip=0pt\relax}
\providecommand{\BIBentryALTinterwordstretchfactor}{4}
\providecommand{\BIBentryALTinterwordspacing}{\spaceskip=\fontdimen2\font plus
\BIBentryALTinterwordstretchfactor\fontdimen3\font minus
  \fontdimen4\font\relax}
\providecommand{\BIBforeignlanguage}[2]{{%
\expandafter\ifx\csname l@#1\endcsname\relax
\typeout{** WARNING: IEEEtran.bst: No hyphenation pattern has been}%
\typeout{** loaded for the language `#1'. Using the pattern for}%
\typeout{** the default language instead.}%
\else
\language=\csname l@#1\endcsname
\fi
#2}}
\providecommand{\BIBdecl}{\relax}
\BIBdecl

\bibitem{abbas2018mobile}
N.~Abbas, Y.~Zhang, A.~Taherkordi, and T.~Skeie, ``Mobile edge computing: A
  survey,'' \emph{IEEE Internet of Things Journal}, vol.~5, no.~1, pp.
  450--465, 2018.

\bibitem{mach2017mobile}
P.~Mach and Z.~Becvar, ``Mobile edge computing: A survey on architecture and
  computation offloading,'' \emph{IEEE Communications Surveys \& Tutorials},
  vol.~19, no.~3, pp. 1628--1656, 2017.

\bibitem{rodrigues2018cloudlets}
T.~G. Rodrigues, K.~Suto, H.~Nishiyama, N.~Kato, and K.~Temma, ``Cloudlets
  activation scheme for scalable mobile edge computing with transmission power
  control and virtual machine migration,'' \emph{IEEE Transactions on
  Computers}, vol.~67, no.~9, pp. 1287--1300, 2018.

\bibitem{zhao2019computation}
J.~Zhao, Q.~Li, Y.~Gong, and K.~Zhang, ``Computation offloading and resource
  allocation for cloud assisted mobile edge computing in vehicular networks,''
  \emph{IEEE Transactions on Vehicular Technology}, vol.~68, no.~8, pp.
  7944--7956, 2019.

\bibitem{you2017energy}
C.~You, K.~Huang, H.~Chae, and B.-H. Kim, ``Energy-efficient resource
  allocation for mobile-edge computation offloading,'' \emph{IEEE Transactions
  on Wireless Communication}, vol.~16, no.~3, pp. 1397--1411, 2017.

\bibitem{8611210}
T.~T. {Nguyen}, L.~{Le}, and Q.~{Le-Trung}, ``Computation offloading in mimo
  based mobile edge computing systems under perfect and imperfect csi
  estimation,'' \emph{IEEE Transactions on Services Computing}, pp. 1--1, 2019.

\bibitem{dai2018joint}
Y.~Dai, D.~Xu, S.~Maharjan, and Y.~Zhang, ``Joint computation offloading and
  user association in multi-task mobile edge computing,'' \emph{IEEE
  Transactions on Vehicular Technology}, vol.~67, no.~12, pp. 12\,313--12\,325,
  2018.

\bibitem{lu2020edge}
H.~Lu, X.~He, M.~Du, X.~Ruan, Y.~Sun, and K.~Wang, ``Edge qoe: Computation
  offloading with deep reinforcement learning for internet of things,''
  \emph{IEEE Internet of Things Journal}, 2020.

\bibitem{yueyuedai}
Y.~Dai, D.~Xu, S.~Maharjan, G.~Qiao, and Y.~Zhang, ``Artificial intelligence
  empowered edge computing and caching for internet of vehicles,'' \emph{IEEE
  Wireless Communications}, vol.~26, no.~3, pp. 12--18, 2019.

\bibitem{rodrigues2019machine}
T.~K. Rodrigues, K.~Suto, H.~Nishiyama, J.~Liu, and N.~Kato, ``Machine learning
  meets computation and communication control in evolving edge and cloud:
  Challenges and future perspective,'' \emph{IEEE Communications Surveys \&
  Tutorials}, 2019.

\bibitem{zhang2019edge}
K.~Zhang, Y.~Zhu, S.~Maharjan, and Y.~Zhang, ``Edge intelligence and blockchain
  empowered 5g beyond for the industrial internet of things,'' \emph{IEEE
  Network}, vol.~33, no.~5, pp. 12--19, 2019.

\bibitem{guo2018mobile}
H.~Guo, J.~Liu, J.~Zhang, W.~Sun, and N.~Kato, ``Mobile-edge computation
  offloading for ultradense iot networks,'' \emph{IEEE Internet of Things
  Journal}, vol.~5, no.~6, pp. 4977--4988, 2018.

\bibitem{rodrigues2016hybrid}
T.~G. Rodrigues, K.~Suto, H.~Nishiyama, and N.~Kato, ``Hybrid method for
  minimizing service delay in edge cloud computing through vm migration and
  transmission power control,'' \emph{IEEE Transactions on Computers}, vol.~66,
  no.~5, pp. 810--819, 2016.

\bibitem{zhao2019novel}
Z.~Zhao, R.~Zhao, J.~Xia, X.~Lei, D.~Li, C.~Yuen, and L.~Fan, ``A novel
  framework of three-hierarchical offloading optimization for mec in industrial
  iot networks,'' \emph{IEEE Transactions on Industrial Informatics}, vol.~16,
  pp. 5424--5434, 2020.

\bibitem{wang2018multi}
F.~Wang, J.~Xu, and Z.~Ding, ``Multi-antenna noma for computation offloading in
  multiuser mobile edge computing systems,'' \emph{IEEE Transactions on
  Communications}, vol.~67, no.~3, pp. 2450--2463, 2018.

\bibitem{samanta2019adaptive}
A.~Samanta and Z.~Chang, ``Adaptive service offloading for revenue maximization
  in mobile edge computing with delay-constraint,'' \emph{IEEE Internet of
  Things Journal}, vol.~6, no.~2, pp. 3864--3872, 2019.

\bibitem{ahmed2019deep}
K.~Ahmed, H.~Tabassum, and E.~Hossain, ``Deep learning for radio resource
  allocation in multi-cell networks,'' \emph{IEEE Network}, 2019.

\bibitem{dai2020deep}
Y.~Dai, D.~Xu, K.~Zhang, S.~Maharjan, and Y.~Zhang, ``Deep reinforcement
  learning and permissioned blockchain for content caching in vehicular edge
  computing and networks,'' \emph{IEEE Transactions on Vehicular Technology},
  vol.~69, no.~4, pp. 4312--4324, 2020.

\bibitem{lu2020blockchain}
Y.~Lu, X.~Huang, K.~Zhang, S.~Maharjan, and Y.~Zhang, ``Blockchain empowered
  asynchronous federated learning for secure data sharing in internet of
  vehicles,'' \emph{IEEE Transactions on Vehicular Technology}, vol.~69, no.~4,
  pp. 4298--4311, 2020.

\bibitem{wang2019smart}
J.~Wang, L.~Zhao, J.~Liu, and N.~Kato, ``Smart resource allocation for mobile
  edge computing: A deep reinforcement learning approach,'' \emph{IEEE
  Transactions on emerging topics in computing}, 2019.

\bibitem{huang2019deep}
L.~Huang, X.~Feng, C.~Zhang, L.~Qian, and Y.~Wu, ``Deep reinforcement
  learning-based joint task offloading and bandwidth allocation for multi-user
  mobile edge computing,'' \emph{Digital Communications and Networks}, vol.~5,
  no.~1, pp. 10--17, 2019.

\bibitem{yang2020lessons}
B.~Yang, X.~Cao, X.~Li, C.~Yuen, and L.~Qian, ``Lessons learned from accident
  of autonomous vehicle testing: An edge learning-aided offloading framework,''
  \emph{IEEE Wireless Communications Letters}, 2020.

\bibitem{zhang2019deep}
K.~Zhang, Y.~Zhu, S.~Leng, Y.~He, S.~Maharjan, and Y.~Zhang, ``Deep learning
  empowered task offloading for mobile edge computing in urban informatics,''
  \emph{IEEE Internet of Things Journal}, 2019.

\bibitem{8690980}
X.~{Chen}, H.~{Zhang}, C.~{Wu}, S.~{Mao}, Y.~{Ji}, and M.~{Bennis},
  ``Performance optimization in mobile-edge computing via deep reinforcement
  learning,'' in \emph{2018 IEEE 88th Veh. Technol. Conf. (VTC-Fall)}, Aug
  2018, pp. 1--6.

\bibitem{xie2019backscatter}
Y.~Xie, Z.~Xu, J.~Xu, S.~Gong, and Y.~Wang, ``Backscatter-aided hybrid data
  offloading for mobile edge computing via deep reinforcement learning,'' in
  \emph{International Conference on Machine Learning and Intelligent
  Communications}.\hskip 1em plus 0.5em minus 0.4em\relax Springer, 2019, pp.
  525--537.

\bibitem{lu2019blockchain}
Y.~Lu, X.~Huang, Y.~Dai, S.~Maharjan, and Y.~Zhang, ``Blockchain and federated
  learning for privacy-preserved data sharing in industrial iot,'' \emph{IEEE
  Transactions on Industrial Informatics}, vol.~16, no.~6, pp. 4177--4186,
  2019.

\bibitem{8399837}
J.~{Xiao}, C.~{Yang}, A.~{Anpalagan}, Q.~{Ni}, and M.~{Guizani}, ``Joint
  interference management in ultra-dense small-cell networks: A multi-domain
  coordination perspective,'' \emph{IEEE Transations on Communication},
  vol.~66, no.~11, pp. 5470--5481, Nov 2018.

\bibitem{Garey:1990:CIG:574848}
M.~R. Garey and D.~S. Johnson, \emph{Computers and Intractability; A Guide to
  the Theory of NP-Completeness}.\hskip 1em plus 0.5em minus 0.4em\relax New
  York, NY, USA: W. H. Freeman \& Co., 1990.

\bibitem{8377343}
J.~{Li}, H.~{Gao}, T.~{Lv}, and Y.~{Lu}, ``Deep reinforcement learning based
  computation offloading and resource allocation for mec,'' in \emph{2018 IEEE
  Wirel. Commun. and Netw. Conf. (WCNC)}, April 2018, pp. 1--6.

\bibitem{lillicrap2015continuous}
T.~P. Lillicrap, J.~J. Hunt, A.~Pritzel, N.~Heess, T.~Erez, Y.~Tassa,
  D.~Silver, and D.~Wierstra, ``Continuous control with deep reinforcement
  learning,'' \emph{arXiv preprint arXiv:1509.02971}, 2015.

\bibitem{qiu2019deep}
C.~Qiu, Y.~Hu, Y.~Chen, and B.~Zeng, ``Deep deterministic policy gradient
  (ddpg)-based energy harvesting wireless communications,'' \emph{IEEE Internet
  of Things Journal}, vol.~6, no.~5, pp. 8577--8588, 2019.

\bibitem{shmoys1993approximation}
D.~B. Shmoys and {\'E}.~Tardos, ``An approximation algorithm for the
  generalized assignment problem,'' \emph{Math. Program.}, vol.~62, no. 1-3,
  pp. 461--474, 1993.

\bibitem{kuhn1955hungarian}
H.~W. Kuhn, ``The hungarian method for the assignment problem,'' \emph{Naval
  Research Logistics (NRL)}, vol.~2, no. 1-2, pp. 83--97, 1955.

\bibitem{ikuno20103gpp}
J.~Ikuno, M.~Wrulich, and M.~Rupp, ``3gpp tr 36.814 v9. 0.0-evolved universal
  terrestrial radio access (e-utra); further advancements for e-utra physical
  layer aspects,'' \emph{Technique Report}, 2010.

\bibitem{wittmann2016chip}
M.~Wittmann, G.~Hager, T.~Zeiser, J.~Treibig, and G.~Wellein, ``Chip-level and
  multi-node analysis of energy-optimized lattice boltzmann cfd simulations,''
  \emph{Concurrency and Computation: Practice and Experience}, vol.~28, no.~7,
  pp. 2295--2315, 2016.

\end{thebibliography}

\end{document}